\documentclass[lettersize,journal]{IEEEtran}
\usepackage{amsmath,amsfonts}
\usepackage{algorithmic}
\usepackage{algorithm}
\usepackage{array}
\usepackage[caption=false,font=normalsize,labelfont=sf,textfont=sf]{subfig}
\usepackage{textcomp}
\usepackage{stfloats}
\usepackage{url}
\usepackage{verbatim}
\usepackage{graphicx}
\usepackage{cite}
\usepackage[colorlinks,
linkcolor=blue,
anchorcolor=blue,
citecolor=blue,backref]{hyperref}
\usepackage[all]{hypcap}
\usepackage{bbding}
\usepackage{multirow}
\usepackage{ulem}
\usepackage{float}
\usepackage{soul, color, xcolor}

\begin{document}
	
	\title{DREMnet: An Interpretable Denoising Framework for Semi-Airborne Transient Electromagnetic Signal}
	
	\author{Shuang Wang, Ming Guo, Xuben Wang, Fei Deng, Lifeng Mao, Bin Wang and Wenlong Gao\vspace{-5mm}    
		\thanks{This work was supported by the Deep Earth National Science and Technology Major Project under Grant 2024ZD1002905 and National Key Research and Development Program of China under Grant 2023YFB3905004. (Corresponding author: Xuben Wang.)}
		\thanks{Shuang Wang, Ming Guo, X. Wang, L. Mao and W. Gao are with the Key Laboratory of Earth Exploration and Information Techniques, Ministry of Education, College of Geophysics, Chengdu University of Technology, Chengdu 610059, China (e-mail: wangs@stu.cdut.edu.cn; mingguo\_cdut@163.com; wxb@cdut.edu.cn; maolifeng07@cdut.cn; gaowenlong@stu.cdut.edu.cn).
			
			Fei Deng, and Bin Wang are with the College of Computer Science and Cyber Security, Chengdu University of Technology, Chengdu 610059, China (e-mail: dengfei@cdut.edu.cn; woldier@foxmail.com).}}
	
	\markboth{Journal of \LaTeX\ Class Files,~Vol.~14, No.~8, August~2021}%
	{Shell \MakeLowercase{\textit{et al.}}: A Sample Article Using IEEEtran.cls for IEEE Journals}
	
	
	\maketitle
	
	\begin{abstract}
	The semi-airborne transient electromagnetic method (SATEM) is capable of conducting rapid surveys over large-scale and hard-to-reach areas. However, the acquired signals are often contaminated by complex noise, which can compromise the accuracy of subsequent inversion interpretations. Traditional denoising techniques primarily rely on parameter selection strategies, which are insufficient for processing field data in noisy environments. With the advent of deep learning, various neural networks have been employed for SATEM signal denoising. However, existing deep learning methods typically use single-mapping learning approaches that struggle to effectively separate signal from noise. These methods capture only partial information and lack interpretability. To overcome these limitations, we propose an interpretable decoupled representation learning framework, termed DREMnet, that disentangles data into content and context factors, enabling robust and interpretable denoising in complex conditions. To address the limitations of convolutional neural network (CNN) and transformer architectures, we utilize the receptance weighted key value (RWKV) architecture for data processing and introduce the contextual weighted key value (Co-WKV) mechanism, which allows unidirectional weighted key value (WKV) to perform bidirectional signal modeling. Our proposed Cover Embedding technique retains the strong local perception of convolutional networks through stacked embedding. Experimental results on test datasets demonstrate that the DREMnet method outperforms existing techniques, with processed field data that more accurately reflects the theoretical signal, offering improved identification of subsurface electrical structures.
	\end{abstract}
	
	
	\section{Introduction}
	\IEEEPARstart{t}{he} semi-airborne transient electromagnetic method (SATEM) is an emerging electromagnetic exploration technique in which the transmitter is deployed on the ground, while the receiver is mounted on an airborne platform \cite{chen2020characteristic}. By performing inversion on the received secondary field signals, it enables accurate extraction of subsurface electrical structures \cite{wu2019development}. SATEM is highly effective for surveying large and inaccessible areas and has been widely applied in mineral resource exploration, groundwater investigation, as well as environmental protection and geological hazard assessment \cite{elliott1996new,mogi2009grounded,ji2013development,ito2011further}.
	
	Since the system conducts observations while in motion, significant noise is inevitably introduced. In SATEM data, typical noise can be divided into natural electromagnetic noise, cultural electromagnetic noise, motion-induced noise \cite{li2017electromagnetic,ji2018noise}, and platform noise \cite{liu2017application}. Natural noise typically manifests as short-term pulses, primarily caused by thunder and lightning \cite{ji2016noising}. Cultural electromagnetic noise refers to stable power line noise and very low-frequency (VLF) radio interference encountered during observations \cite{wu2022denoising}. In some application scenarios, intense pulse interference may occur, such as ore concentration areas \cite{macnae1984comparison}. Motion-induced noise is a non-stationary noise generated as the sensor moves with the airborne platform, caused by irregularities in the geomagnetic field. Platform noise originates from electronic components onboard the airborne platform, such as propulsion and communication systems. Compared to other noise sources, motion-induced noise is characterized by large amplitude, wide distribution, and low frequency \cite{macnae1984noise,mccracken1986minimization}. Additionally, due to the long observation distance, the received signal strength is relatively weak. The combination of weak signals and complex noise presents substantial challenges in SATEM data processing, making noise suppression and removal crucial steps in the process.
	
	Current SATEM data denoising methods can be broadly divided into two categories: traditional filtering techniques and deep learning-based approaches. Traditional filter-based methods primarily rely on empirical parameter settings, which are determined by observing the physical characteristics of the acquired data. For instance, Lv \textit{et al.} \cite{lv2022noise} proposed a denoising scheme combining wavelet threshold filtering with adaptive singular value decomposition filtering. Sun \textit{et al.} \cite{sun2022motion} introduced a simple yet effective single-period polynomial fitting method to remove motion noise from secondary field data. Yang \textit{et al.} \cite{yang2022motion} developed an overdetermined linear system for late-time motion noise in semi-airborne transient electromagnetic data, based on Fourier series, which was then solved using least-squares inversion for denoising. Ma \textit{et al.} \cite{ma2023receiver} derived a calculation formula for mutual inductance coupling between the transmitter and receiver, and subsequently proposed a hybrid correction scheme using a two-stage strategy to correct receiver noise. Most traditional methods rely heavily on the selection of empirical parameters, which requires significant expertise from data processors \cite{wu2019removal}. Moreover, these methods are generally more suitable for data with low noise levels and often struggle to effectively suppress complex, high-noise field data.
	\begin{figure*}[!t]
		\centering
		\includegraphics[width=6in]{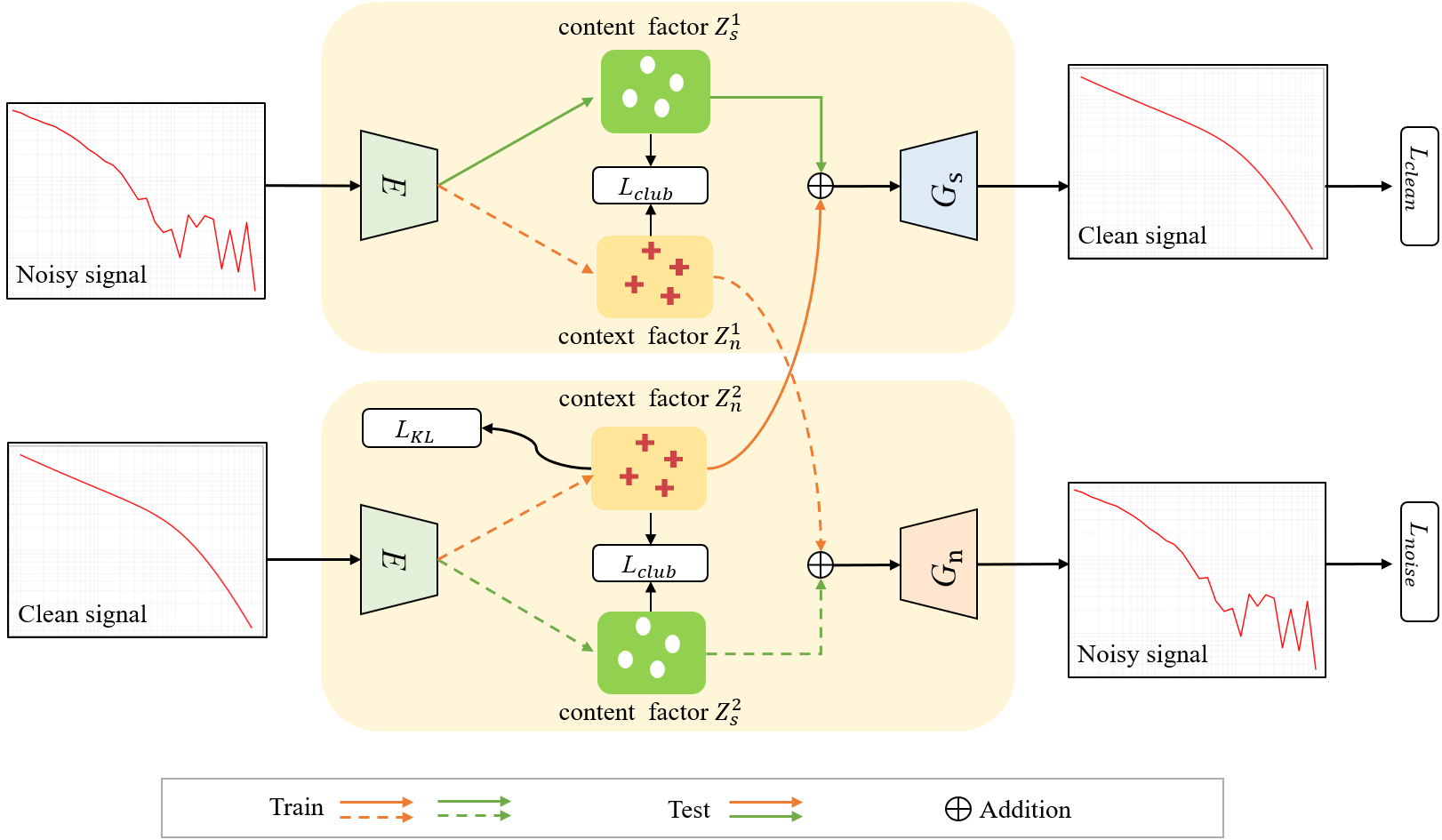}
		\caption{The overall framework of DREMnet, the data is encoded by the encoder \( E \) into content factors and context factors. The decoder $G_s$ is employed to reconstruct the accurate clean signal, while the decoder $G_n$ is used to ensure the correctness of the disentangled representations.}
		\label{1}
	\end{figure*}
	
	Deep learning-based methods exploit the powerful learning and modeling capabilities of neural networks to perform denoising, offering advantages such as end-to-end processing, high speed, and strong noise reduction performance \cite{zhang2017learning,potluri2011cnn,zhang2019deep,long2015fully}. For example, Wu \textit{et al.} \cite{wu2019removal} applied a wavelet neural network to remove motion noise in helicopter-borne transient electromagnetic data. Chen \textit{et al.} \cite{chen2020temdnet} innovatively transformed one-dimensional signals into two-dimensional sequences and applied an image denoising network to transient electromagnetic data, achieving promising results. Wu \textit{et al.} \cite{wu2021noising} combined long short-term memory (LSTM) networks with autoencoder structures for transient electromagnetic data denoising. Li \textit{et al.} \cite{li2023inceptcn} proposed a novel method called IncepTCN-SISC, which integrates deep learning and dictionary learning techniques—inception-temporal convolutional networks and shift-invariant sparse coding—to suppress strong noise in controlled-source electromagnetic (CSEM) data. Additionally, Li \textit{et al.} \cite{li2024gtcn} introduced a gated temporal convolutional network (GTCN) to map noisy CSEM sequences to high-quality outputs. Pan \textit{et al.} \cite{pan2023tem1dformer} designed a one-dimensional denoising network based on the encoder structure of the vision transformer \cite{dosovitskiy2020image}, effectively reducing noise in transient electromagnetic data. Deng \textit{et al.} \cite{deng2024semi} proposed a denoising method using variation diffusion model (VDM), which introduced constraint conditions to guide the diffusion model \cite{ho2020denoising} and achieve the desired denoising results. Cheng \textit{et al.} \cite{cheng2025pc} utilized parallel convolution to extract multi-scale features from input data and employed a bidirectional LSTM (BiLSTM) to establish complex mappings between features and clean data, enabling effective denoising of transient electromagnetic signals. Lin \textit{et al.} \cite{lin2024semi} introduced a recurrent self-coding neural network (RSCN) specifically designed for one-stop noise suppression along an entire survey line.
	
	The aforementioned methods rely on predicting either noise or signal, representing a single-mapping learning approach. This singular representation often struggles to effectively separate signal from noise \cite{liang2023interpretable}, resulting in representations that contain both components. Consequently, these methods face challenges in isolating noise and signal when processing complex field data and can only capture partial information from the signal. Furthermore, convolutional network structures, due to their inherent limitations \cite{10681481}, are insufficient for global modeling of time-series signals. In contrast, Transformer and VDM methods are computationally intensive, with their memory and computational complexity increasing quadratically with sequence length, rendering standard Transformers impractical and costly for long-sequence signals. Additionally, existing solutions are typically end-to-end mapping approaches, which lack interpretability.
		\begin{figure}[!t]
		\centering
		\includegraphics[width=3.5in]{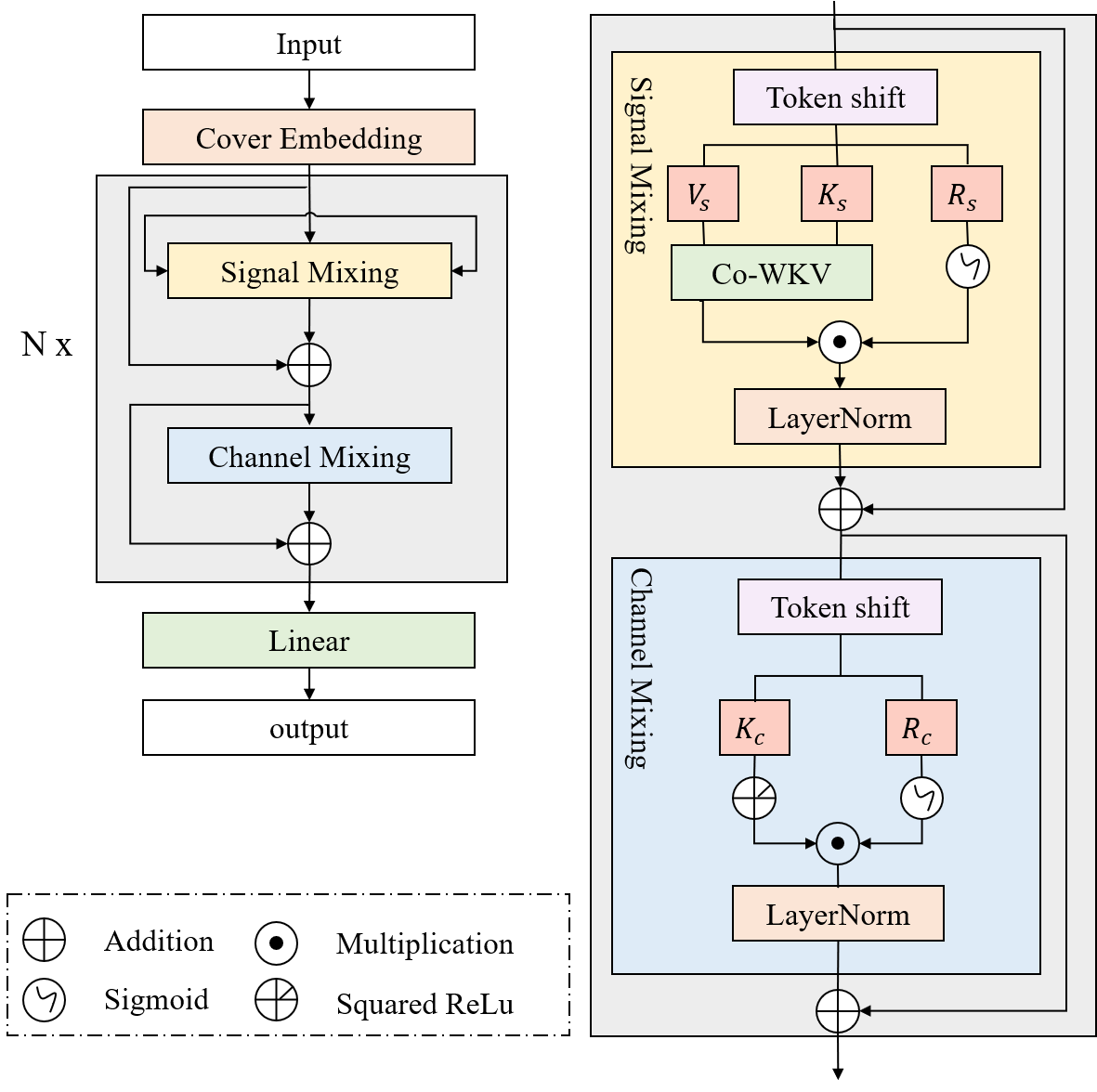}
		\caption{On the left is the architecture of the encoder $E$ and decoders $G_s$ and $G_n$ used in DREMnet; on the right is the structure of the core module DR block within the encoder and decoders, which consists of a signal mixing module and a channel mixing module.}
		\label{2}
	\end{figure}
	To address the aforementioned issues and enhance interpretability, we propose an interpretable disentangled representation learning framework, termed DREMnet. This framework separates the data into content and context factors, enabling robust denoising of SATEM data under complex conditions while improving interpretability. To overcome the limitations of convolutional neural network (CNN) and Transformer architectures, we adopt the Receptance Weighted Key Value (RWKV) \cite{hou2024rwkv,peng2024eagle} architecture for data processing. The original RWKV’s weighted key value (WKV) \cite{peng2023rwkv} attention mechanism is unidirectional; however, for SATEM signals, the current signal is correlated with both preceding and succeeding signals. To facilitate bidirectional modeling, we introduce the contextual weighted key value (Co-WKV) mechanism, which extends the unidirectional WKV to support bidirectional signal modeling. To preserve the strong local perception capability of convolutional networks, we propose the Cover Embedding technique, which incorporates overlapping embeddings to ensure each token retains information directly relevant to it. Experimental results on the test set demonstrate that the DREMnet method achieves superior denoising performance. Furthermore, inversion results from processed field data indicate that the DREMnet-processed data closely align with theoretical signals, providing a more accurate reflection of subsurface features.

	The contributions of this paper are as follows:
	
	1. We propose an interpretable disentangled representation learning framework that decomposes data into content and context factors, enabling robust and interpretable denoising of SATEM data under complex conditions.
	
	2. The proposed Co-WKV mechanism extends the unidirectional WKV to support bidirectional signal modeling. Additionally, we introduce Cover Embedding, which overlays embeddings to ensure each token retains information directly relevant to it. This approach effectively mitigates the limitations of CNN and Transformer architectures while preserving their advantages.

	\section{Method}
	\subsection{Framework Overview}
	The observed SATEM signal is typically represented as the sum of two components: noise and the response signal:
	\begin{equation}
		y = s + n
	\end{equation}
	where \( y \) is the observed signal, \( s \) is the response signal, and \( n \) represents the noise. Most deep learning denoising methods focus on single representation learning, where the neural network predicts either the noise or the signal. However, residual learning has been shown to offer superior performance \cite{zhang2017beyond}. As a result, most networks adopt noise prediction to enable denoising, represented as:
	\begin{equation}
		s' = s + n - \text{net}(s+n) 
	\end{equation}
	Where, \(\text{net}(s+n)\) represents the neural network's prediction of the noise in the observed signal \( y\). The training objective of the network is to minimize \( n - \text{net}(s+n) \). This single-representation learning approach exhibits inherent limitations in separating noise from signals. In contrast, decoupled representation learning models critical data representations by disentangling information into task-relevant and task-irrelevant factors \cite{wang2024disentangled}. For denoising applications, the informative signal components are considered task-relevant factors, while the noise components are task-irrelevant factors. Therefore, applying decoupled representation learning to separate data into content and context factors enables robust and interpretable denoising under complex conditions. The overall framework is shown in Fig. ~\ref{1}.
	\begin{figure}[!t]
		\centering
		\includegraphics[width=3in]{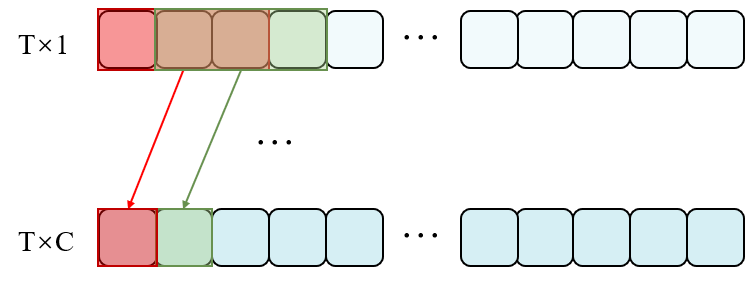}
		\caption{Cover Embedding, as illustrated in the case where the cover length is 3, combines the current signal data along with the next two signal data into a token with a dimension of \( C \).}
		\label{3}
	\end{figure} 
	
	The DREMnet framework consists of three core components: an encoder \( E \), a denoising decoder \( G_s \), and a representation decoder \( G_n \). The encoder \( E \) encodes input data into content and context factors. The denoising decoder \( G_s \) reconstructs clean signals by decoding a combination of content factors extracted from noisy signals and context factors derived from noise-free references. The representation decoder \( G_n \) decodes the combination of content and context factors into the corresponding signals, ensuring the correctness of the decoupled representation.
		
	Given the collected data \( x \), the encoder \( E \) encodes it into a pair of disentangled representations:
	\begin{equation}
		Z_s, Z_n = E(x)
	\end{equation}
	Here, \( Z_s \) represents the content factors, and \( Z_n \) represents the context factors. To ensure the distributions of \( Z_s \) and \( Z_n \) are fully separated, we employ a mutual information upper bound estimator, CLUB \cite{cheng2020club}, since the distributions of the signal and noise are entirely distinct. Intuitively, because the clean signal contains no noise, we treat the distribution of the context factors of the clean signal as a zero vector, \( Z_n = 0 \), and use KL divergence to constrain its distribution.
	\begin{figure*}[!t]
		\centering
		\includegraphics[width=6in]{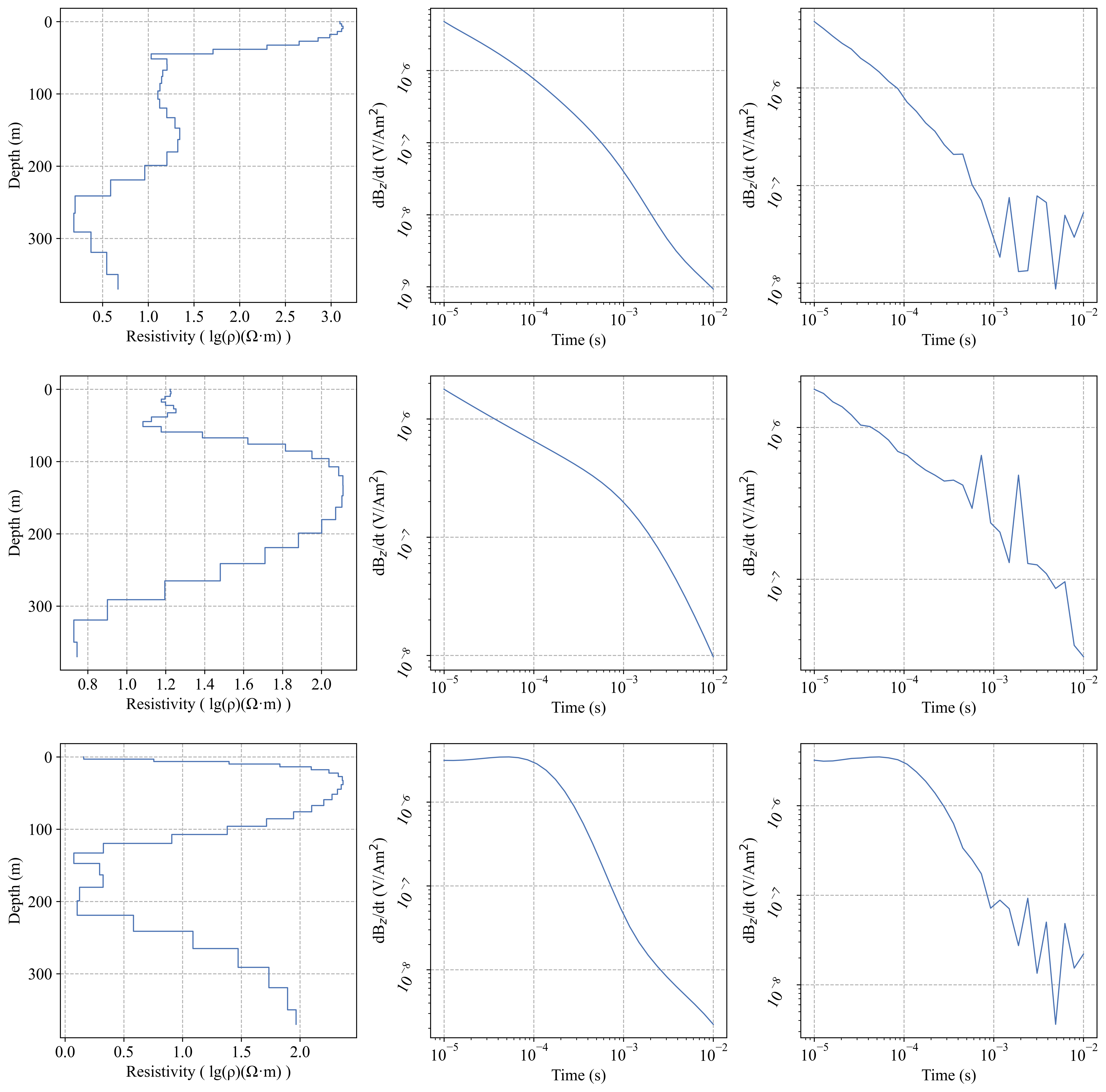}
		\caption{Examples of resistivity models as well as the forward data and the noise-added data. The left displays the resistivity models, the middle shows the forward data and the right presents the noise-added data.}
		\label{4}
	\end{figure*}
		\begin{figure}[!t]
		\centering
		\includegraphics[width=3in]{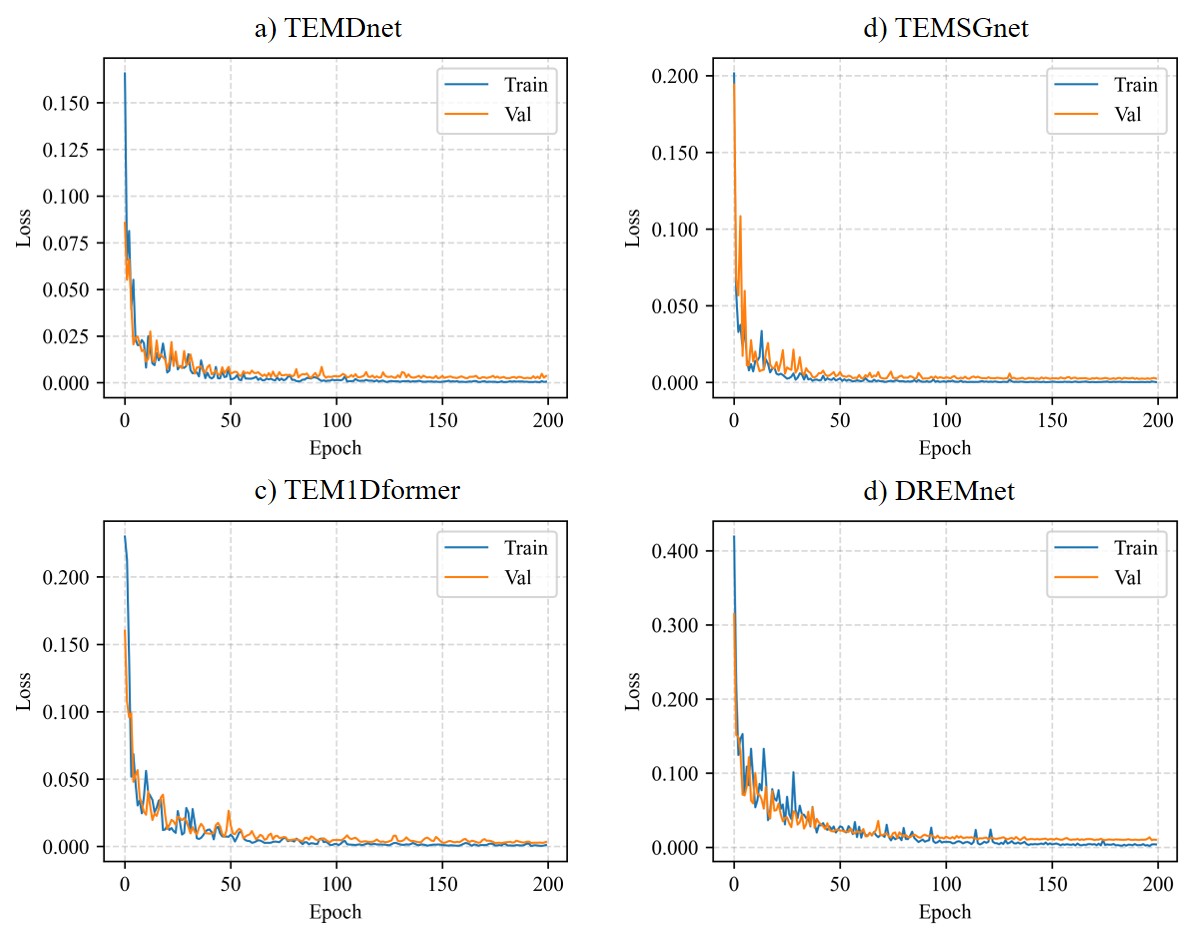}
		\caption{The training and validation loss curves of the four deep learning models.}
		\label{loss}
	\end{figure}
	\begin{figure}[!t]
		\centering
		\includegraphics[width=3in]{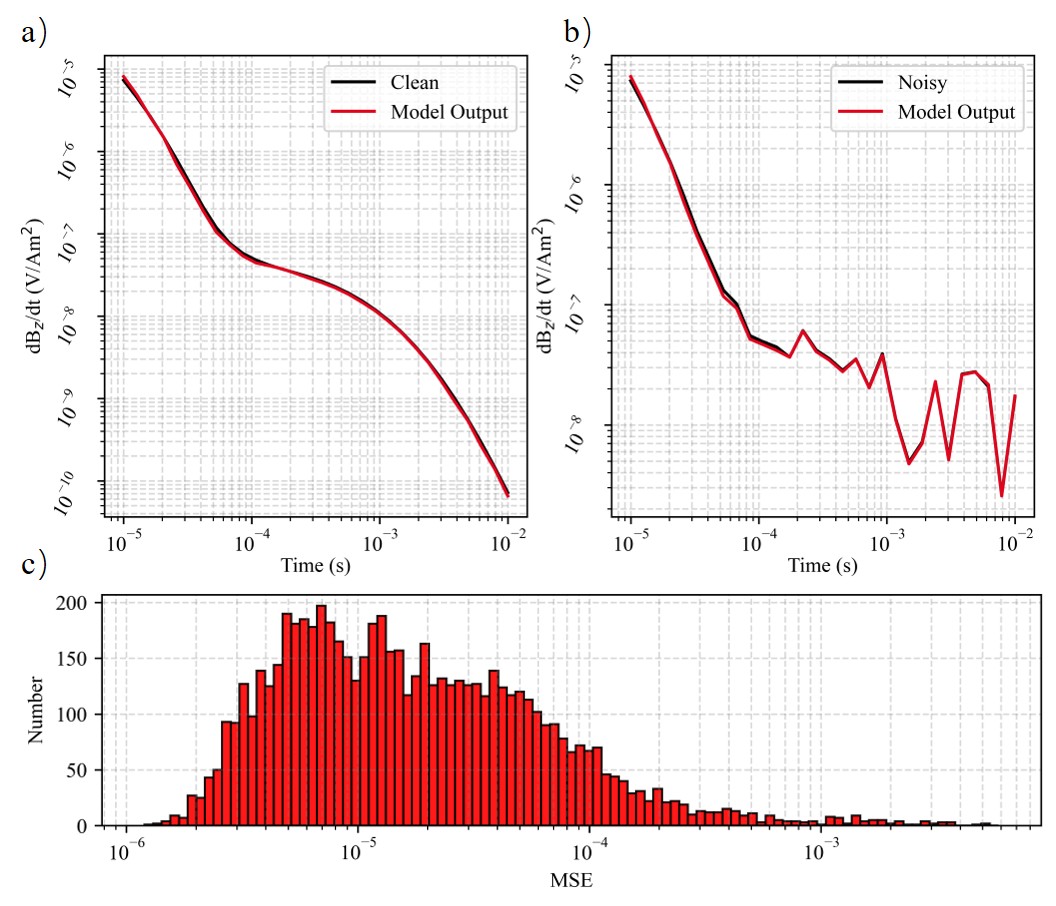}
		\caption{The clean signal \( s \) is disentangled into signal factors \( Z_s^1 \) and noise factors \( Z_n^1 \), while the noisy signal \( n \) is disentangled into signal factors \( Z_s^2 \) and noise factors \( Z_n^2 \). (a) The clean signal is decoded using \( (Z_s^2, Z_n^1) \). (b) The noisy signal is decoded using \( (Z_s^1, Z_n^2) \). (c) The MSE statistics of the decoded data from \( (Z_s^1, Z_n^2) \) on the test set, compared to the noisy signal.}
		\label{5}
	\end{figure}
	
	During training, the encoder \( E \) encodes both the noisy signal \( n \) and the clean signal \( s \). We aim for the network to correctly learn the disentangled representations: \( \{Z_s^1, Z_n^1\} = E(n) \) and \( \{Z_s^2, Z_n^2\} = E(s) \). The content factors \( Z_s^1 \) and \( Z_s^2 \) should follow the same distribution, indicating that the content factor representations are similar, while the context factors should differ, as the clean signal contains no noise. Consequently, we swap the context factors between the two representations and use the representation decoder to decode them. Decoding \( (Z_s^2, Z_n^1) \) should yield the noisy signal, confirming that the data has been fully disentangled. Finally, the denoising decoder \( G_s \) decodes the combination of content and context factors to produce the denoised signal.
	\begin{equation}
		s = G_s(Z_s^1,Z_n^2)
	\end{equation}
	
	The overall loss function is:
	\begin{equation}
		L=L_{clean}+L_{noise}+L_{KL}
	\end{equation}
	\begin{equation}
		L_{clean} = ||G_s(Z_s^1, Z_n^2) - s||^2
	\end{equation}
	\begin{equation}
		L_{noise} = ||G_n(Z_s^2, Z_n^1) - n||^2
	\end{equation}
	\begin{equation}
		L_{KL} = (\mu^2 + \sigma^2 - log(\sigma^2) -1)
	\end{equation}
	where $\mu$ and $\sigma$ denote the mean and standard deviation of \( Z_n^2 \), respectively.
	\subsection{RWKV-Based Encoder and Decoder}
	To overcome the limitations of CNN and Transformer architectures \cite{yang2024restore,duan2024vision}, we employ the RWKV architecture for data processing. Both the encoder and decoder adopt the same architecture, as illustrated in Fig. \ref{2}. After the SATEM signal is input, it first passes through a Cover Embedding module (detailed in the Cover Embedding section) to enhance local signal information. The signal is then processed through \( N \) DR blocks before being passed through a linear layer to produce the output. Our design retains the advantages of RWKV with only necessary modifications.

	The DR blocks comprises a signal mixing module and a channel mixing module. The signal mixing module functions as the attention mechanism, enabling linear global attention calculations. Meanwhile, the channel mixing module facilitates the fusion of token features across channels.
	
	The signal mixing module can perform linear global attention on the signal processed through Cover Embedding \( x \in \mathbb{R}^{T \times C} \). Where, \( T \) represents the signal length, and \( C \) denotes the embedding dimension. The signal mixing module first performs a token shift, following the original design of RWKV without any modifications. The resulting \( x_s \) after the token shift is then processed through three linear layers to obtain \( R_s \), \( K_s \) and \( V_s \).
	\begin{equation}
		R_s=x_sW_r,K_s=x_sW_k,V_s=x_sW_v
	\end{equation}
	Where \( W_r \), \( W_k \) and \( W_v \) represent the weights of the three linear layers, then use Co-WKV (see the Co-WKV subsection) to perform global linear attention calculations, the global attention result $wkv$ can be expressed as:
	\begin{equation}
		wkv=Co-WKV(K_s,V_s)
	\end{equation}
	Finally, the attention results are adjusted using \( R_s \) to produce the output.
	\begin{figure*}[!h]
		\centering
		\includegraphics[width=7in]{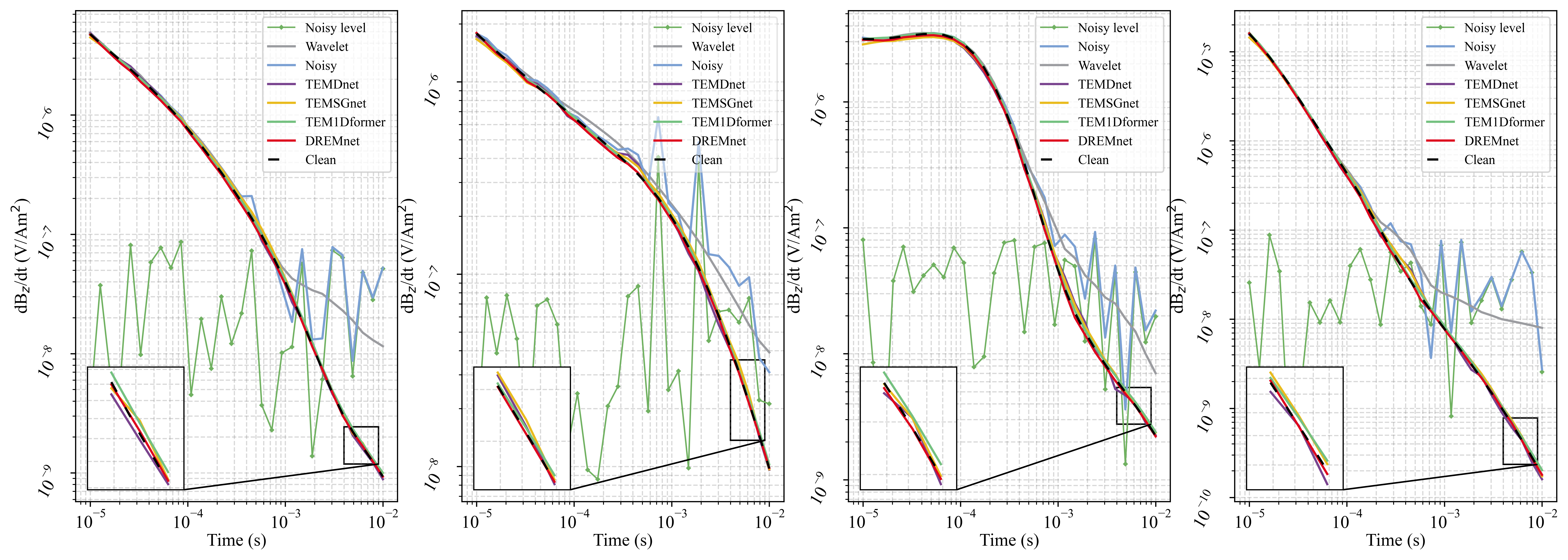}
		\caption{Four examples of denoising results for the five methods on the test dataset.}
		\label{6}
	\end{figure*}
	
	The channel mixing module performs feature fusion on tokens across channels, processing the token sequence after signal mixing, \( x \in \mathbb{R}^{T \times C} \). Channel mixing also applies a token shift, followed by two linear layers to obtain \( R_c \) and \( K_c \), ultimately producing the output through feature fusion.
	\subsection{Co-WKV}
	The original RWKV's WKV attention mechanism is unidirectional, which limits the attention receptive field and does not provide global attention. Therefore, we utilize Co-WKV, an improved WKV attention mechanism, to enable global attention computation. Given the inputs \( K_s \in \mathbb{R}^{T \times C} \) and \( V_s \in \mathbb{R}^{T \times C} \), the attention computation result \( wkv_t \) for the \( t \)-th token can be expressed as:
	\begin{equation}
		wkv_t=\frac{\sum_{i=0,i\neq t}^{T-1}{e^{-\left(\left|t-i\right|-1\right)w+k_i}v_i+e^{u+k_t}v_t}}{\sum_{i=0,i\neq t}^{T-1}{e^{-\left(\left|t-i\right|-1\right)w+k_i}+e^{u+k_t}}}
	\end{equation}
	Where, \( T \) represents the number of tokens, and \( k_i \in \mathbb{R}^C \) and \( v_i \in \mathbb{R}^C \) denote the \( i \)-th token in \( K_s \in \mathbb{R}^{T \times C} \) and \( V_s \in \mathbb{R}^{T \times C} \), respectively. \( w \in \mathbb{R}^C \) and \( u \in \mathbb{R}^C \) are two learnable parameters representing the decay weight of the relative position bias \( -\left(\left|t-i\right|-1\right) \) and the reward for the current \( t \)-th token. Therefore, \( wkv_t \) is essentially the weighted sum of \( v_1 \) to \( v_t \), with the weights determined by \( k_i \), the relative position bias, and the parameters \( w \) and \( u \).
	\subsection{Cover Embedding}
	To ensure that each token contains information relevant to its directly associated data and to endow DREMnet with the strong local perceptual advantages of convolutional networks, we applied Cover Embedding to the input signal. Cover Embedding is a type of overlapping embedding, as illustrated in Fig. \ref{3}. For an input signal of length \( T \) and a cover length of \( n \), Cover Embedding combines the current signal data with the subsequent \( n-1 \) signal data into a token with a dimension of \( C \). Consequently, the final signal of length \( T \) is transformed into a token sequence \( T \times C \) after applying Cover Embedding.
	\begin{figure*}[!t]
		\centering
		\includegraphics[width=5in]{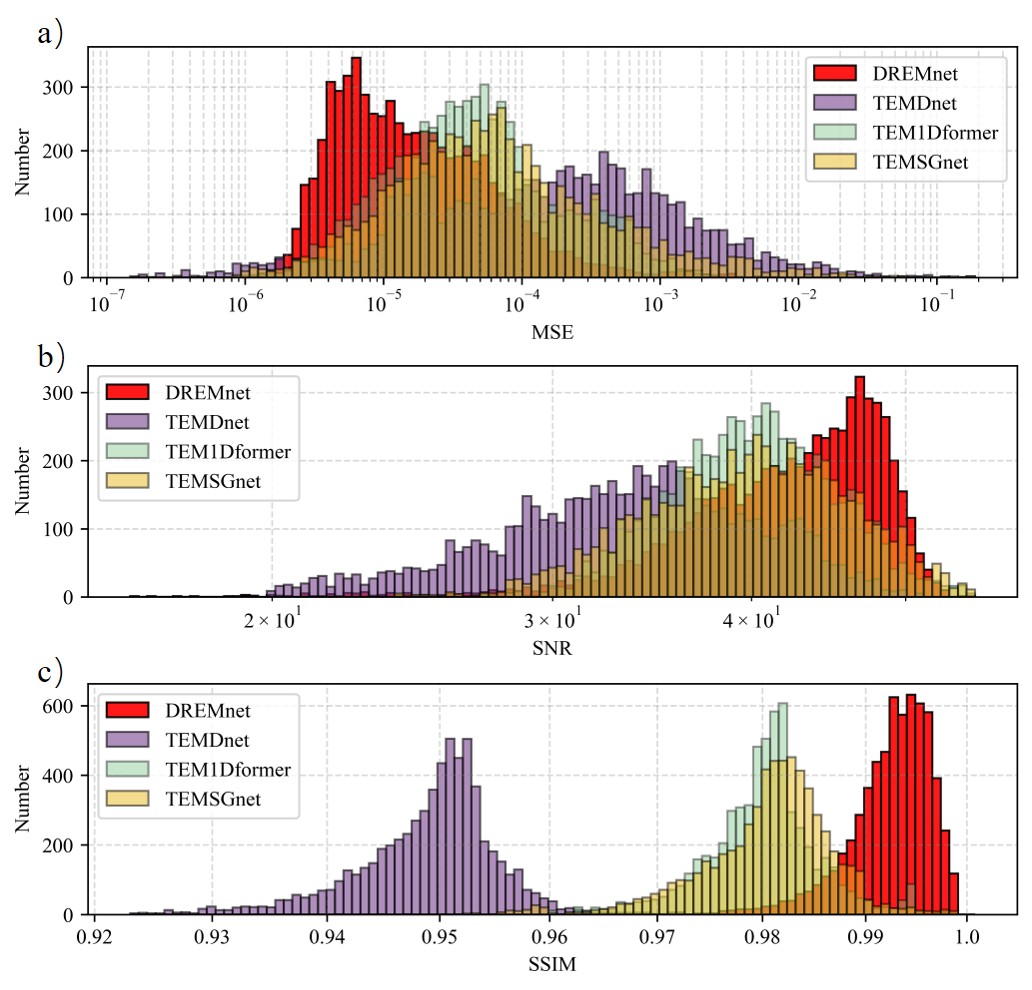}
		\caption{Statistics of quantitative evaluation metrics. (a) MSE statistics of the denoising results across the entire test set. (b) SNR statistics of the denoising results across the entire test set. (c) SSIM statistics of the denoising results across the entire test set.}
		\label{7}
	\end{figure*}
	\section{Experiments}
	\subsection{Evaluation Metrics}
	To quantitatively assess the quality of the denoising results, we selected three commonly used metrics: mean squared error (MSE), signal-to-noise ratio (SNR), and structural similarity index measure (SSIM). MSE measures the error between the denoised results and the ground truth, and its calculation formula is as follows:
	\begin{equation}
		MSE=\frac {1}{n}\sum_{i=1}^{n}(x_{i}^{d}-x_{i}^{t})^2
	\end{equation}
	Where $n$ denotes the number of data points, $x_{i}^{d}$ represents the data denoised by the network, $x_{i}^{t}$ represents the ground truth, and the closer the MSE value is to 0, the closer the denoising result is to the ground truth. SNR measures the quality of the denoising result, calculated by the following formula:
	\begin{equation}
		SNR=10{log}_{10}\frac{{||x_d||}_F^2}{{||x_t-x_d||}_F^2}
	\end{equation}
	Where \( x_d \) represents the data after denoising by the network, \( x_t \) denotes the ground truth, and \( \|\cdot\|_F \) represents the Frobenius norm. A larger SNR value indicates a higher quality of denoising. The SSIM measures the structural consistency of the denoised results, and its calculation formula is as follows:
	\begin{equation}
		SSIM=\frac{(2u_du_t+c_1)(2\sigma_{dt}+c_2)}{(u_d^2{+u}_t^2+c_1)(\sigma_d^2{+\sigma}_t^2+c_2)}
	\end{equation}
	Where \( u_d \) is the mean of the denoised result, \( u_t \) is the mean of the ground truth, \( \sigma_{dt} \) is the covariance between the denoised result and the ground truth, \( \sigma_d \) is the variance of the denoised result, and \( \sigma_t \) is the variance of the ground truth. The constants \( c_1 \) and \( c_2 \) are introduced to avoid numerical instability. A SSIM value closer to 1 indicates that the denoised result is closer to the ground truth.
	\begin{figure}[!h]
		\centering
		\includegraphics[width=3in]{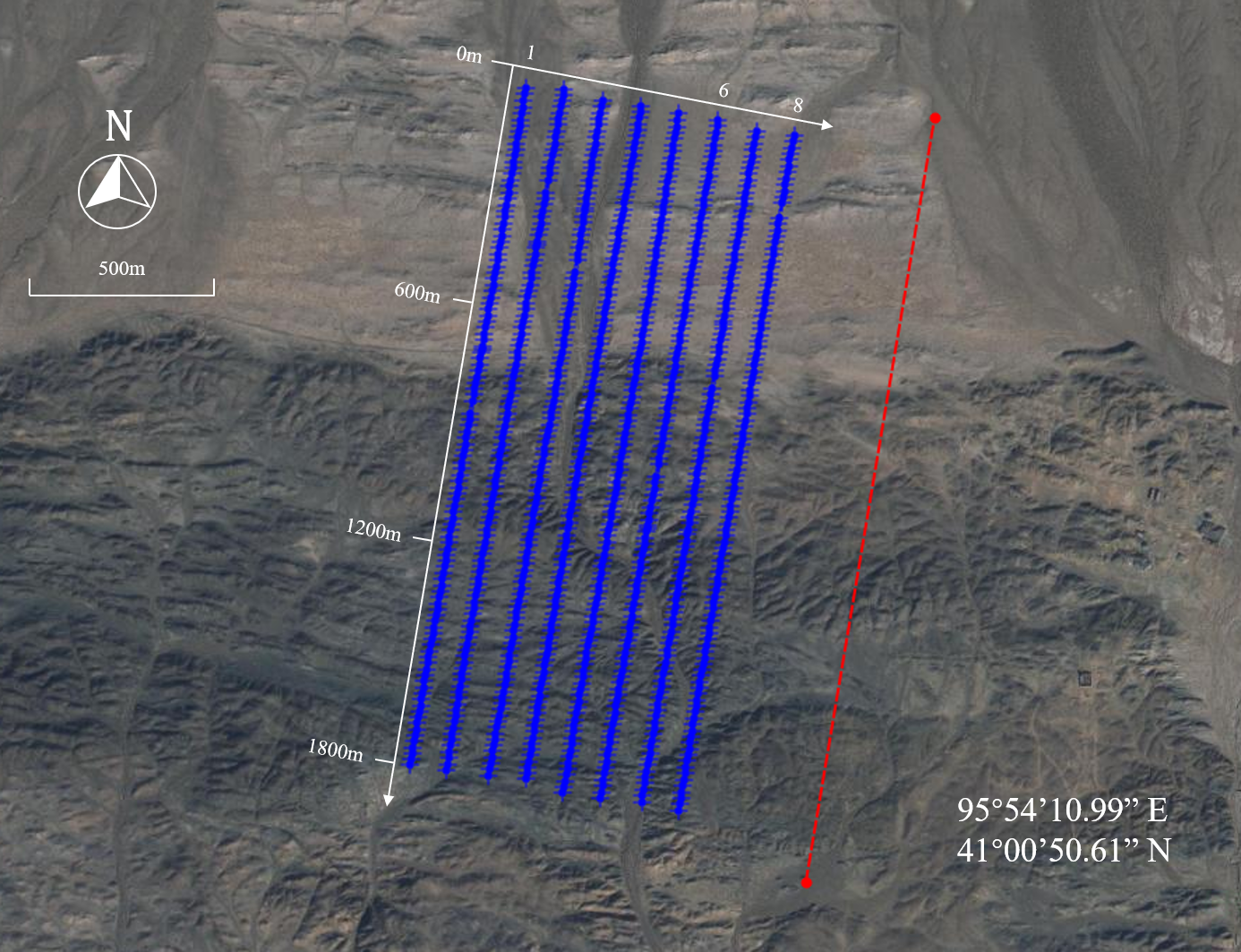}
		\caption{The coordinates of the center point of the line source in the field exploration area are 95°54’10.99”E and 41°00’50.61”N.}
		\label{8}
	\end{figure}
	\subsection{Dataset}
	We utilized the large resistivity model database (RMD) developed by \cite{asif2023dl}, which contains a variety of geologically plausible and geophysically resolvable subsurface structures. The database comprises approximately one million resistivity models, with resistivity values ranging from 1 to 2000 ohm-meters, consisting of 30 layers and a maximum depth of 500 meters. Each model adheres to physical constraints. RMD has been shown to improve performance and generalization while also enhancing the consistency and reliability of deep learning models \cite{liu2024multi}.
	
	We selected 50,000 models from the dataset and conducted one-dimensional forward modeling on them to obtain the corresponding forward responses. Then, we introduced noise into the forward response data. Part of the noise was derived from actual field measurements, which includes motion-induced noise, nearby or moderately distant sferics noise, cultural and natural electromagnetic noise. As a type of incidental impulse noise, the amount of extraction of the nearby or moderately distant sferics noise was relatively small, while another part was Gaussian noise, as complex field noise distributions are often approximated by a Gaussian distribution \cite{chen2020temdnet}. For the remaining noise, we selected a simulation method that has been shown to closely replicate actual noise \cite{auken2008resolution}, which is defined as:
		\begin{equation}
			s_n=s+N(0,1)[STD^2+(\frac{n}{s})^2]^\frac{1}{2}s
		\end{equation}
	Where, \(s_n\) represents the obtained noisy signal, and \(s\) is the forward theoretical data. \(N(0,1)\) denotes the standard Gaussian distribution. STD is the uniform noise, and \(n\) is the background noise contribution, which is defined as:
		\begin{equation}
			n=b{(\frac{t}{{10}^{-3}})}^{-\frac{1}{2}}
		\end{equation}
	Where, \( b \) is the noise level at 1 ms. It is typically taken between 1 nV/m² and 5 nV/m² \cite{auken2008resolution}. We divided the dataset into training, validation, and test sets in a ratio of 8:1:1, resulting in 40,000 samples for training, and 5,000 samples each for validation and testing. For some models, the forward data and the data after noise addition are shown in Fig. \ref{4}.

	\subsection{Train}
	During training, the DR block was set to 12, the batch size was set to 32, and AdamW was used as the optimizer with a learning rate of 0.0001. The training was conducted for 200 epochs on an Nvidia A6000 GPU. 
	
	We selected three methods for comparative analysis: TEMDnet \cite{chen2020temdnet}, TEMSGnet \cite{deng2024semi}, and TEM1Dformer \cite{pan2023tem1dformer}. TEMDnet represents convolutional and fully connected approaches, transforming one-dimensional signals into two-dimensional formats and applying image denoising techniques for noise reduction. The authors of TEMDnet have made their code publicly available on GitHub, which we used for training. TEMSGnet exemplifies generative methods, using noisy signals as conditional inputs to enable the diffusion model to generate clean signals conditionally. We trained TEMSGnet using the code provided by its authors on GitHub. TEM1Dformer embodies transformer-based methods that utilize attention mechanisms, employing the vision transformer (ViT) encoder structure in its network architecture. Although the authors have not released their code, we replicated their approach based on ViT's code structure and use the settings outlined in their paper during training. Specifically, the encoder consists of 24 layers with a hidden dimension of 512, eight attention heads with a head dimension of 64, and an MLP width of 2048. Throughout the training process, we ensured that all methods utilized the same dataset and same training costs. All other basic settings were kept completely consistent as well. Additionally, we selected a traditional method employing wavelet thresholding for a more comprehensive comparison. The wavelet basis function was set to Sym6, the number of decomposition levels to four, and the threshold was estimated using the soft-thresholding estimation function \cite{peng2000noising}.
	
	Fig. \ref{loss} shows the loss curves of the four models during training. It can be observed that all four models have converged as the training progresses.
	
	The training time, GPU memory consumption during training, inference time, and inference throughput of the four models are summarized in Table~\ref{tab:train}. Since TEMSGnet is based on a diffusion model, it requires iterative sampling during inference, resulting in the lowest inference throughput. DREMnet, on the other hand, involves multiple loss functions, leading to slightly longer training times compared to TEM1Dformer. However, after training, DREMnet demonstrates superior performance in inference efficiency.
	
	\begin{figure*}[!t]
		\centering
		\includegraphics[width=6.5in]{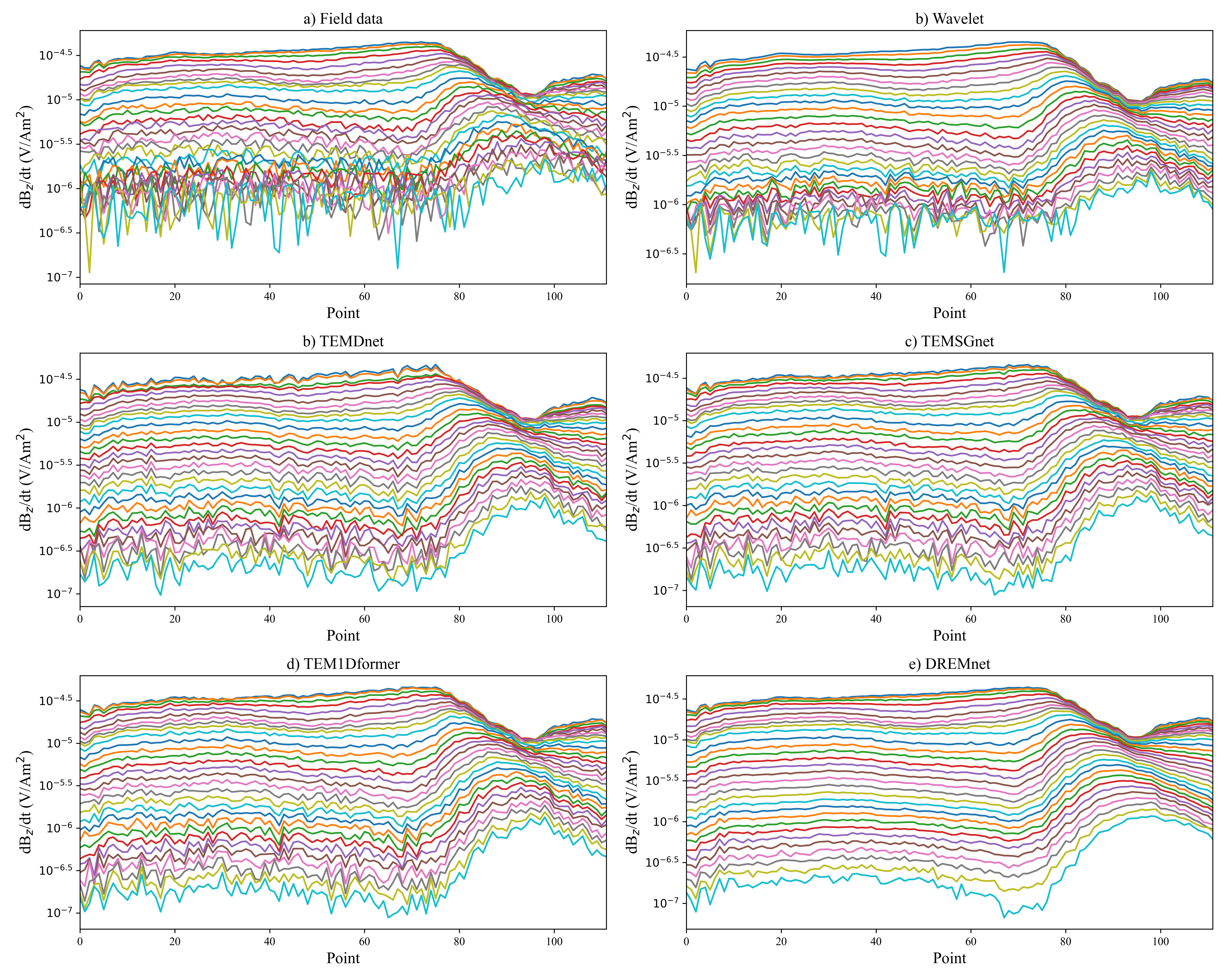}
		\caption{The denoising results for survey line 6 are as follows: (a) represents the field data, (b) is the denoising result from wavelet thresholding, (c) is the denoising result from TEMDnet, (d) is the denoising result from TEMSGnet, (e) is the denoising result from TEM1Dformer, and e is the denoising result from DREMnet.}
		\label{9}
	\end{figure*}
		\begin{table*}[t]
		\caption{Comparison of efficiency metrics.}\label{tab:train}
		\centering
		\linespread{1.25} \selectfont
		\begin{tabular}{cccccc}
			\hline
			& Epoch Time (S) & Training Time (H) & GPU Memory (GB)  & Denoise Time (S) (100 Signals) & Throughput  \\ \hline
			Wavelet&   \textbackslash       & \textbackslash    & \textbackslash & 0.1 & \textbackslash \\
			TEMDnet&    50.3   & 2.79 & 3.6 & 0.15 & 3792\\
			TEMSGnet&    58.7   & 3.26 & 6.3 & 0.75 & 1711\\
			TEM1Dformer&    65.2   & 3.61 & 5.1 & 0.23 & 2748\\
			DREMnet&    71.8   & 3.98 & 4.2 & 0.17 & 3012\\ \hline
		\end{tabular}
	\end{table*}
	\subsection{Test}
	\subsubsection{Synthetic data}
		\begin{table}[!t]
		\caption{Comparison of Five Denoising Methods on the Simulation Dataset.}\label{tab:1}
		\centering
		\linespread{1.25} \selectfont
		\begin{tabular}{ccccc}
			\hline
			&                 & MSE       & SNR     & SSIM   \\ \hline
			& Noisy           & 0.3761 & 13.7893 & 0.7075 \\
			& Wavelet         & 0.0642 & 21.7304 & 0.8656 \\
			& TEMDnet         & 3.9748e-4 & 36.0067 & 0.9499 \\
			& TEMSGnet        & 9.6898e-5 & 43.7438 & 0.9810 \\
			& TEM1Dformer     & 8.5346e-5 & 44.2069 & 0.9823 \\
			& DREMnet         & \pmb{1.8061e-5} & \pmb{47.4325} & \pmb{0.9959} \\ \hline
		\end{tabular}
	\end{table}
	To verify that the encoder \( E \) correctly disentangles the data into content and context factors, we analyzed a set of clean and noisy signals and conducted a statistical analysis on the test set, as shown in Fig. \ref{5}. The clean signal \( s \) is disentangled into content factors \( Z_s^1 \) and context factors \( Z_n^1 \), while the noisy signal \( n \) is disentangled into content factors \( Z_s^2 \) and context factors \( Z_n^2 \). Decoding \( (Z_s^2, Z_n^1) \) with \( G_s \), as shown in Fig. \ref{5}(a), accurately reconstructs the clean signal. Similarly, decoding \( (Z_s^1, Z_n^2) \) with \( G_n \), as illustrated in Fig. \ref{5}(b), successfully reconstructs the noisy signal. These results demonstrate that the model can effectively disentangle the data. Additionally, we computed the MSE histogram between the decoded data from \( (Z_s^1, Z_n^2) \) and the noisy signal across the test set, as shown in Fig. \ref{5}(c). The results indicate that the network exhibits stable disentanglement performance, successfully separating data into content and context factors.
	\begin{figure*}[!t]
		\centering
		\includegraphics[width=7in]{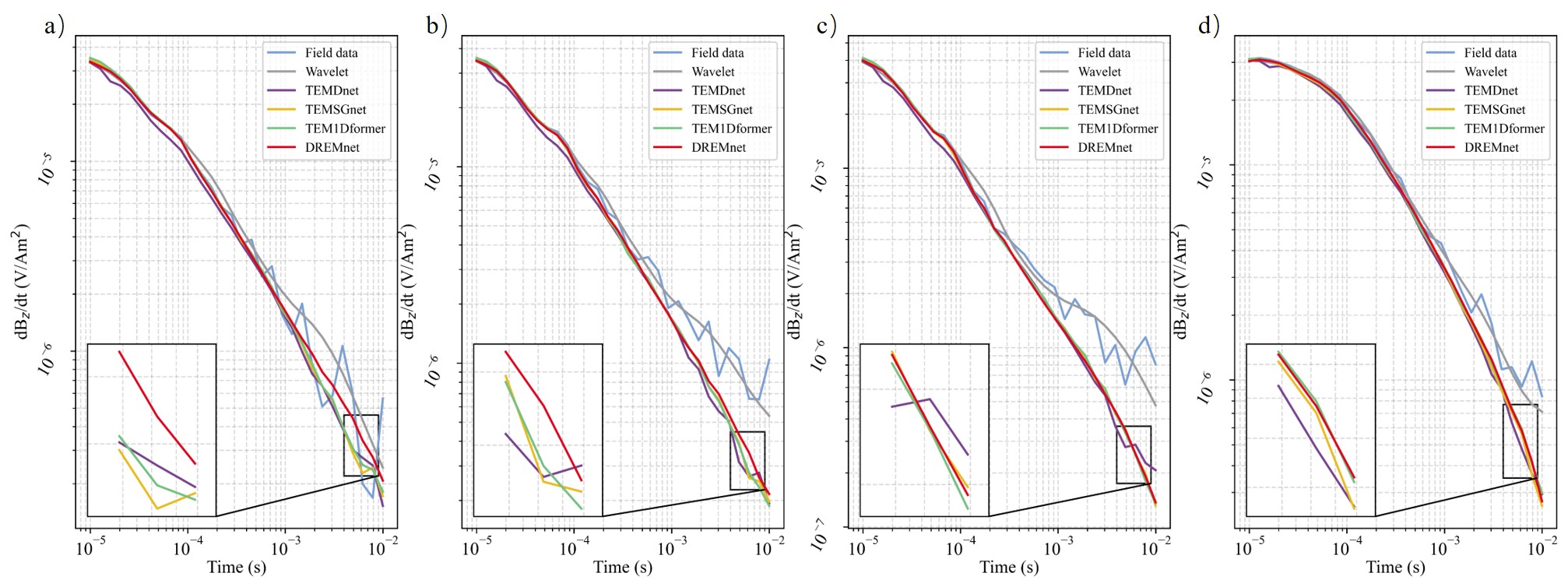}
		\caption{Comparison of denoising results for different measurement points on survey line 6. (a) corresponds to measurement point 20, (b) corresponds to measurement point 40, (c) corresponds to measurement point 60, (d) corresponds to measurement point 80.}
		\label{10}
	\end{figure*}
		\begin{figure*}[!t]
		\centering
		\includegraphics[width=7in]{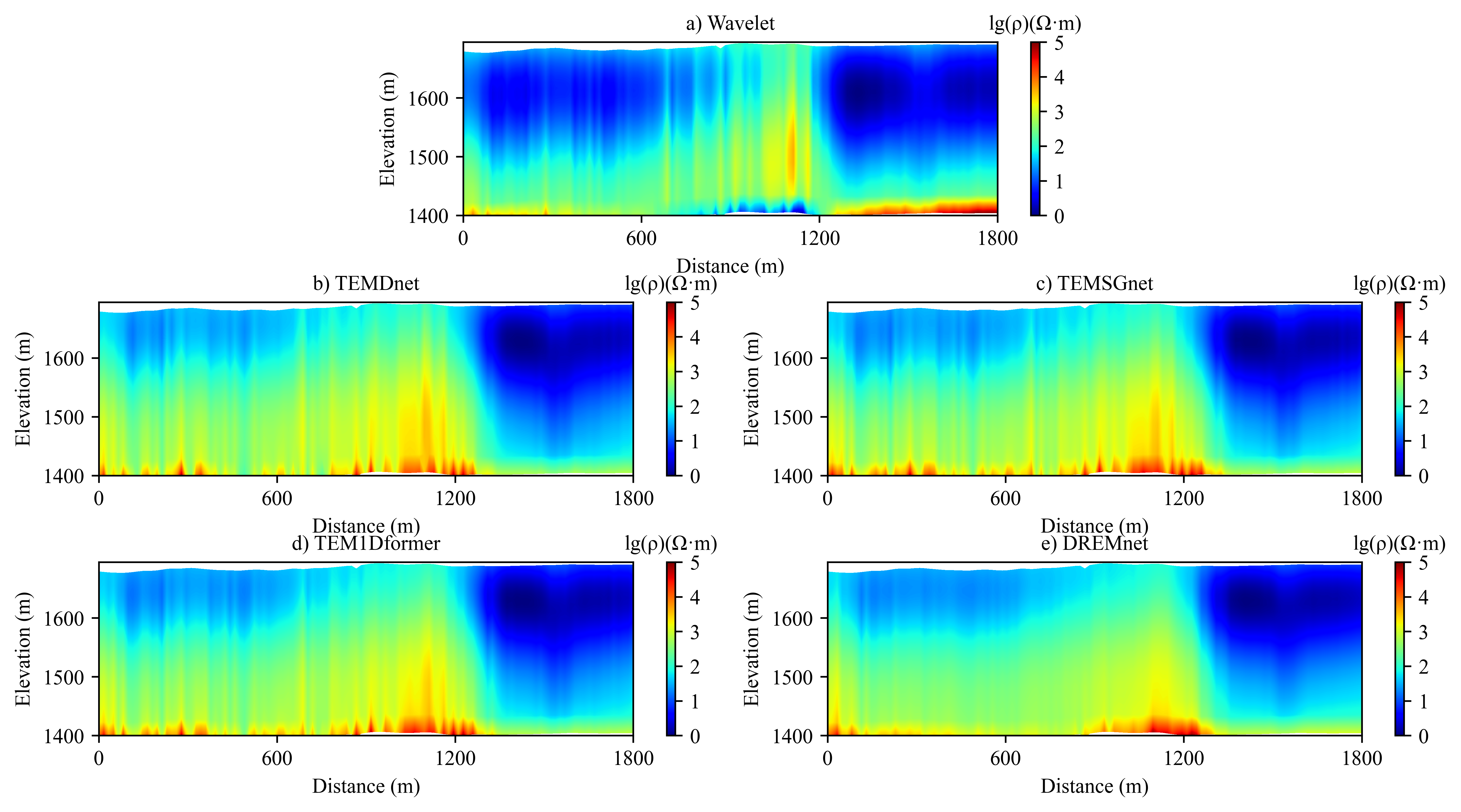}
		\caption{The inversion results of survey line 6: (a) wavelet thresholding denoising result, (b) TEMDnet denoising result, (c) TEMSGnet denoising result, (d) TEM1Dformer denoising result, (e) DREMnet denoising result.}
		\label{11}
	\end{figure*}
	
	To evaluate the effectiveness of our method, we conducted tests on the test set, and the results are presented in Fig. \ref{6}. As shown, our method closely aligns with the theoretical clean data. In contrast, TEMDnet, based on convolution, and TEMSGnet, which uses a generative model, exhibit the largest deviation from the theoretical data due to their limited temporal awareness. Although TEM1Dformer, a transformer-based model, demonstrates improved temporal perception, its performance is slightly inferior to that of our proposed method.
	
	To quantitatively evaluate the performance of the five methods, we calculated the average of three evaluation metrics, as shown in Table~\ref{tab:1}. The results indicate that our method's denoising outcome is the closest to the theoretical signal, achieving the highest SNR. In contrast, the CNN-based method performed the worst, likely due to the inherent modeling limitations of convolutional networks.
	
	To evaluate the stability of the proposed method, we conducted a statistical analysis on the entire test set, as shown in Fig. \ref{7}. Fig. \ref{7} (a) presents the MSE histogram, where the results of DREMnet are concentrated around \( 1 \times 10^{-5} \), indicating smaller values compared to other methods. This suggests that DREMnet offers superior denoising performance with stable results. Fig. \ref{7} (b) displays the SNR histogram, where DREMnet's results are concentrated around 45, achieving a higher SNR than the other methods.
	\subsubsection{Field data}
	\begin{figure*}[!t]
		\centering
		\includegraphics[width=6in]{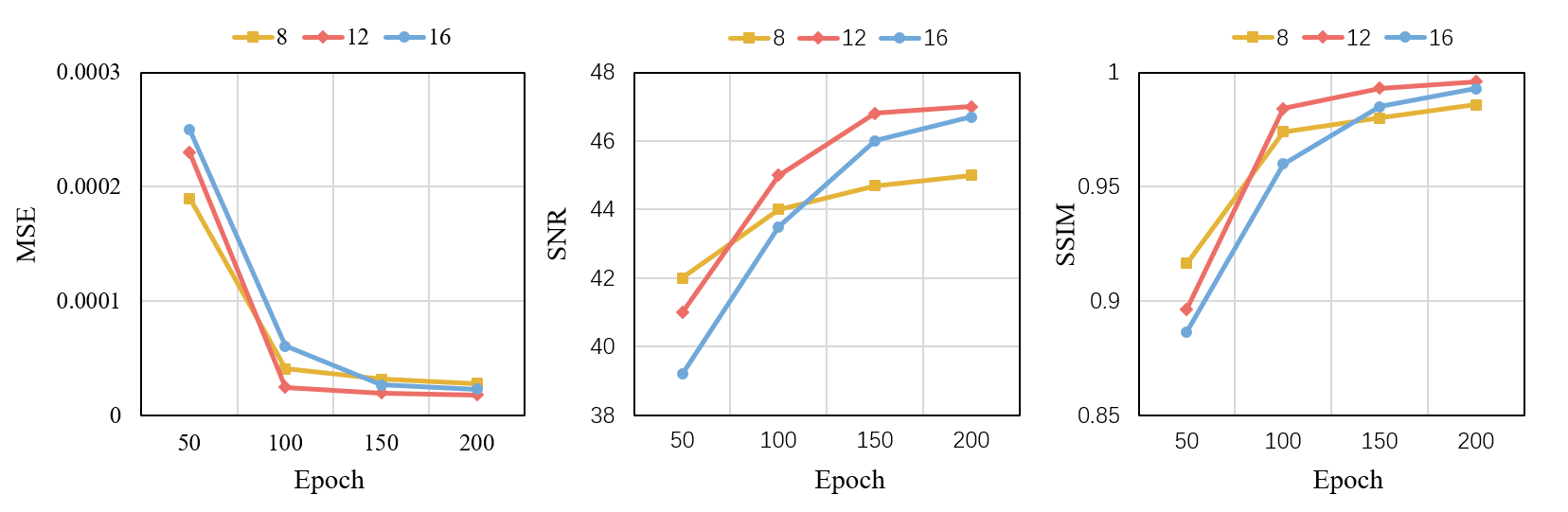}
		\caption{Ablation study. Changes in three evaluation metrics during training with the number of DR blocks set to 8, 12 and 16, respectively.}
		\label{12}
	\end{figure*}
	To evaluate the model's denoising capability on field signals, we conducted a denoising experiment using field data collected from a survey area in Gansu, China, as shown in Fig. \ref{8}. The blue lines in the figure represent the survey lines, while the red line is the line source with a length of 2000 m. The transmitter model used is TXU30. The receiver operated at a flight altitude of 30 m, with a flight speed of 5 m/s and a sampling frequency of 192 kHz.

	We directly applied the model trained on synthetic data to denoise the field data. Fig. \ref{9} presents the denoising results for survey line 6. As shown, although TEMDnet and TEMSGnet successfully reduced cross-talk in some lower channels, they were unable to resolve it in the upper channels, and some noise remained in the lower channels. TEM1Dformer also showed room for improvement in denoising the lower channels. In contrast, DREMnet demonstrated the best performance, effectively eliminating cross-talk across all channels, providing clear signal contours, and significantly reducing noise in the lower channels, resulting in a smoother attenuation curve.

	Fig. \ref{10} presents the denoising results at the 20th, 40th, 60th, and 80th measurement points along survey line 6. As shown, the results processed by DREMnet exhibit a smoother attenuation curve, even in the later stages. In contrast, the denoising results from other neural networks still show some oscillations in the later stages of the curve.

	To further evaluate the denoising performance of the methods, we performed inversion \cite{lu2022quasi} on the denoised results. The inversion results are shown in Fig. \ref{11}. As observed, the results processed by DREMnet more accurately reflect the geoelectric structure, while the inversion results of other methods display relatively blurred boundaries. In the high-resistivity region on the right side of the figure, DREMnet's results exhibit higher resolution, clearer boundary information, and a more continuous transverse direction, indicating a higher SNR in the data processed by DREMnet.
	\subsection{Ablation Study}
	To evaluate the reasonableness of the number of DR blocks $N$, we conducted ablation study by varying $N$ to 8, 12 and 16, respectively. During training, we recorded the changes in three evaluation metrics. As shown in Fig. \ref{12}, when $N = 8$, the model capacity is insufficient to fully represent the data, resulting in degraded performance. In contrast, when $N = 12$ and $N = 16$, increasing the model capacity does not lead to significant performance improvements, indicating that 12 is a reasonable choice. Furthermore, when $N = 16$, the model converges more slowly, likely due to the increased training cost required for a larger model to reach optimal performance.
	
	We then conducted ablation experiments on the Co-WKV, Cover Embedding, and disentanglement mechanism to quantify their respective contributions. When the disentanglement mechanism was removed, it was replaced with a single mapping scheme. As shown in Table~\ref{tab:ablation}, each component contributes positively to model performance, and the best results are achieved when all components are used together.
	\begin{table}[!t]
		\caption{Ablation Study.}\label{tab:ablation}
		\centering
		\linespread{1.25} \selectfont
		\begin{tabular}{p{0.065\textwidth} p{0.07\textwidth} p{0.085\textwidth} p{0.065\textwidth} p{0.05\textwidth} p{0.05\textwidth}}
			\hline
		Co-WKV	&    Cover Embedding &Disentanglement Mechanism & MSE       & SNR     & SSIM   \\ \hline
		$\times$&\checkmark	& \checkmark           & 9.6856e-5 & 43.4587 & 0.9824 \\
		\checkmark&	$\times$& \checkmark         & 8.2456e-5 & 44.7436 & 0.9832 \\
	\checkmark	&\checkmark	& $\times$         & 1.3658e-4 & 41.2731 & 0.9786 \\
	\checkmark	&\checkmark	& \checkmark         & \pmb{1.8061e-5} & \pmb{47.4325} & \pmb{0.9959} \\ \hline
		\end{tabular}
	\end{table}
	
	\section{Conclusions}
	This paper presents an interpretable framework for SATEM signal denoising based on disentangled representation learning. By leveraging this approach, the data is decomposed into content and context factors, facilitating robust and interpretable denoising of SATEM signals under complex conditions. The proposed method utilizes the RWKV architecture, overcoming the limitations of CNNs and transformers while retaining their advantages. Experiments on synthetic datasets demonstrate that the method outperforms others in denoising performance and effectively denoises field data using a model trained on synthetic data. The inversion results of the processed field data show that the proposed method yields results closer to the theoretical signal, offering a more accurate representation of subsurface features.
	
	DREMnet is based on disentangled representation learning to achieve interpretable denoising of SATEM data under complex conditions. Experimental results demonstrate that our model offers improved interpretability, robustness, and applicability. However, several limitations remain. First, the proposed approach is still supervised learning, meaning the model’s performance and generalization heavily depend on the quality of the dataset. For example, a model trained on noise data from survey area A can effectively process data from survey area A but may experience performance degradation when applied to survey area B, where noise levels differ significantly. Second, the network requires multiple loss functions to constrain each component, resulting in slower convergence during training. Future work will explore incorporating contrastive learning to reduce reliance on labeled data from new survey area.
\normalem
\bibliography{r}

\begin{IEEEbiography}[{\includegraphics[width=1in,height=1in,clip,keepaspectratio]{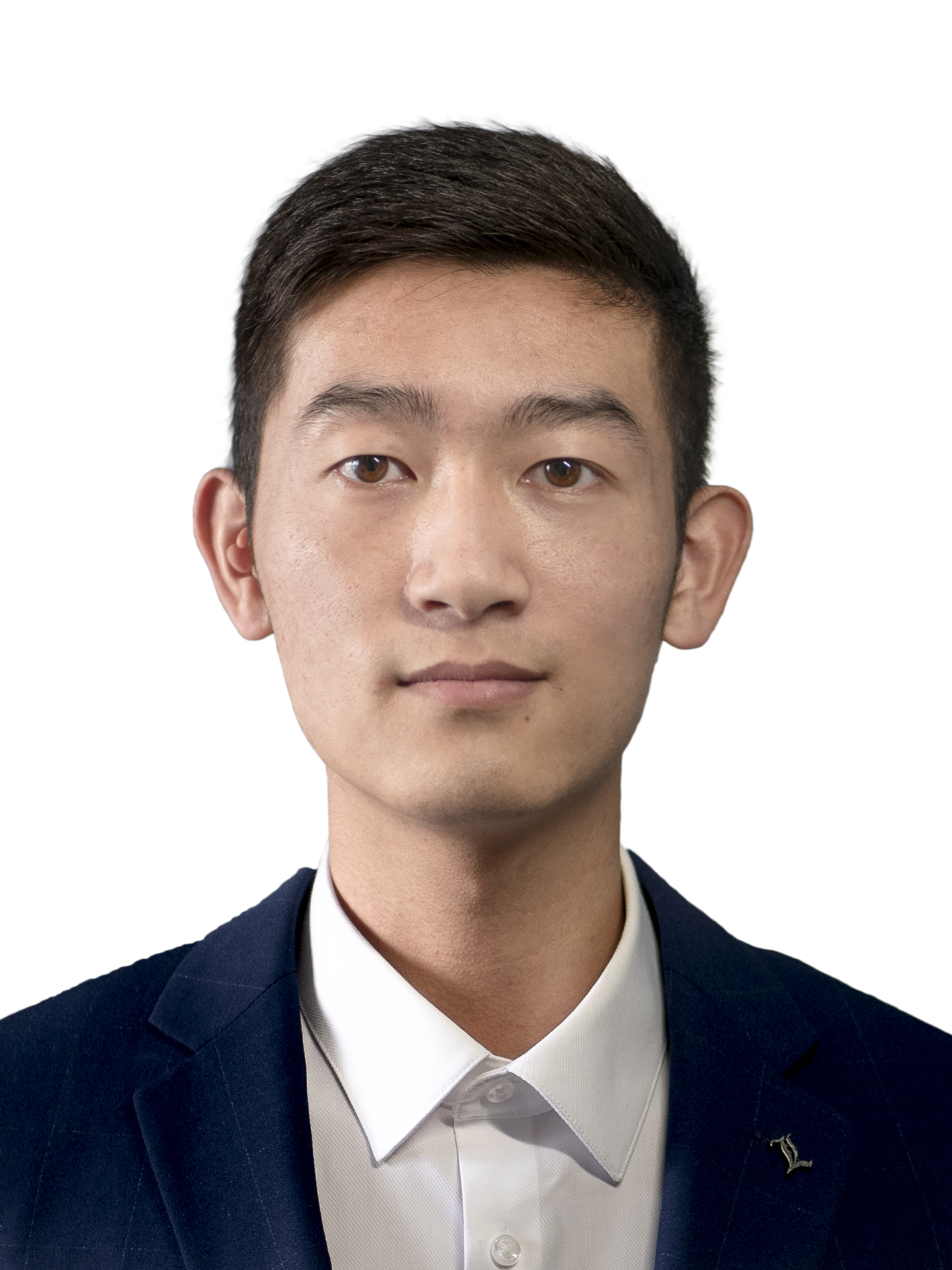}}]
	{Shuang Wang} received the B.S. degree in software engineering from the Chengdu University of Technology, Chengdu, China, in 2022. He is currently pursuing his Ph.D degree in Earth Exploration and Information Technology at Chengdu University of Technology, Chengdu, China. His research interests include applications of deep learning, computer vision, complex signal processing, intelligent geophysical data processing and modeling.
\end{IEEEbiography}
\vspace{-12pt}
\begin{IEEEbiography}[{\includegraphics[width=1in,height=1in,clip,keepaspectratio]{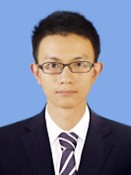}}]
	{Ming Guo} received the B.S. degree in geophysics from the Chengdu University of Technology, Chengdu, China, in 2017, and the M.Sc. degree in earth exploration and information technology from the from Chengdu University of Technology, Chengdu, China, in 2021. He is currently pursuing the Ph.D. degree with Chengdu University of Technology, Chengdu, China. His research interests include the data processing and inversion methods of electromagnetic geophysics
\end{IEEEbiography}
\vspace{-12pt}
\begin{IEEEbiography}[{\includegraphics[width=1in,height=1in,clip,keepaspectratio]{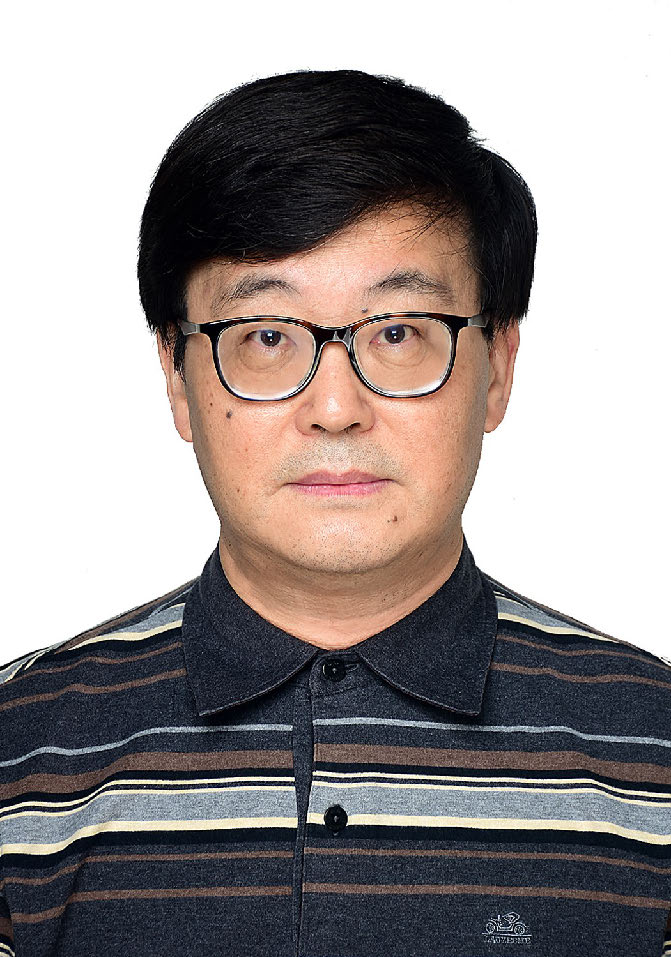}}]
	{Xuben Wang} received the Ph.D. degree in geophysics from the College of Geophysics, Chengdu University of Technology, Chengdu, China, in 2002. His main research interests include magnetotellurics and airborne electromagnetic forward modeling and inversion theory as well as the multiphysics interpretation of continental dynamics.
\end{IEEEbiography}
\vspace{-12pt}
\begin{IEEEbiography}[{\includegraphics[width=1in,height=1in,clip,keepaspectratio]{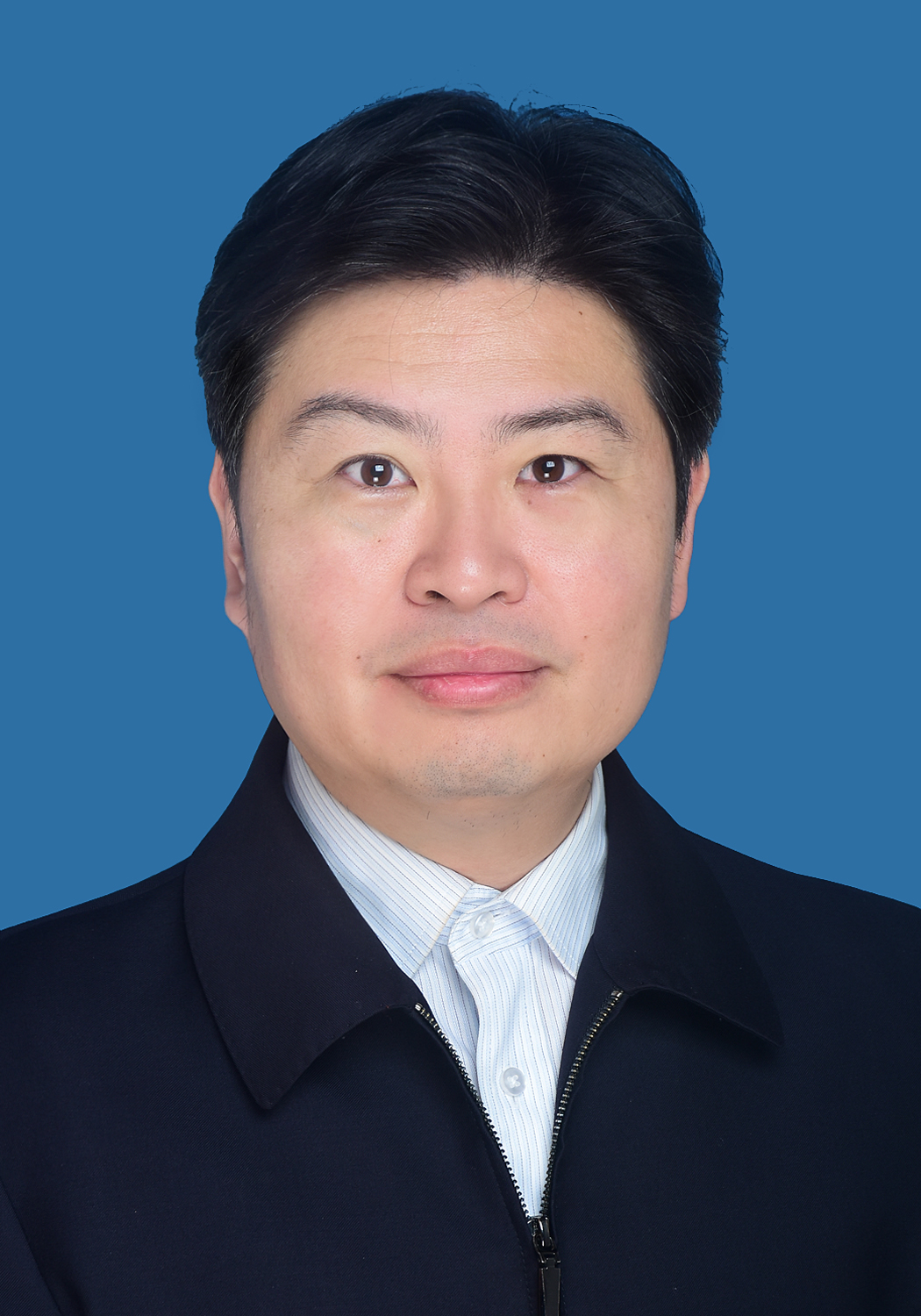}}]
	{Fei Deng} has been with the College of Computer and Network Security, Chengdu University of Technology, where he is currently a Professor. His research interests include geophysical three-dimensional modeling methods, deep learning, and computer graphics.
\end{IEEEbiography}
\vspace{-12pt}
\begin{IEEEbiography}[{\includegraphics[width=1in,height=1in,clip,keepaspectratio]{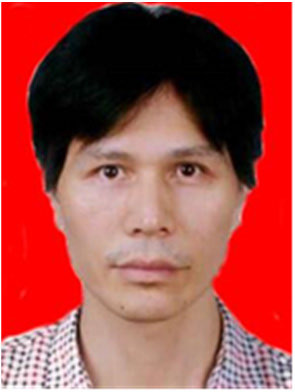}}]
	{Lifeng Mao} received the M.Sc. and Ph.D. degrees in geophysics from Chengdu University of Technology, Chengdu, China, in 2004 and 2007, respectively. He is currently an Associate Professor with the College of Geophysics, CDUT. His research interests include the forward and inversion theory of geophysical electromagnetic methods, with a focus on the 3-D simulation of electromagnetic fields in both time and frequency domains.
\end{IEEEbiography}
\vspace{-12pt}
\begin{IEEEbiography}[{\includegraphics[width=1in,height=1in,clip,keepaspectratio]{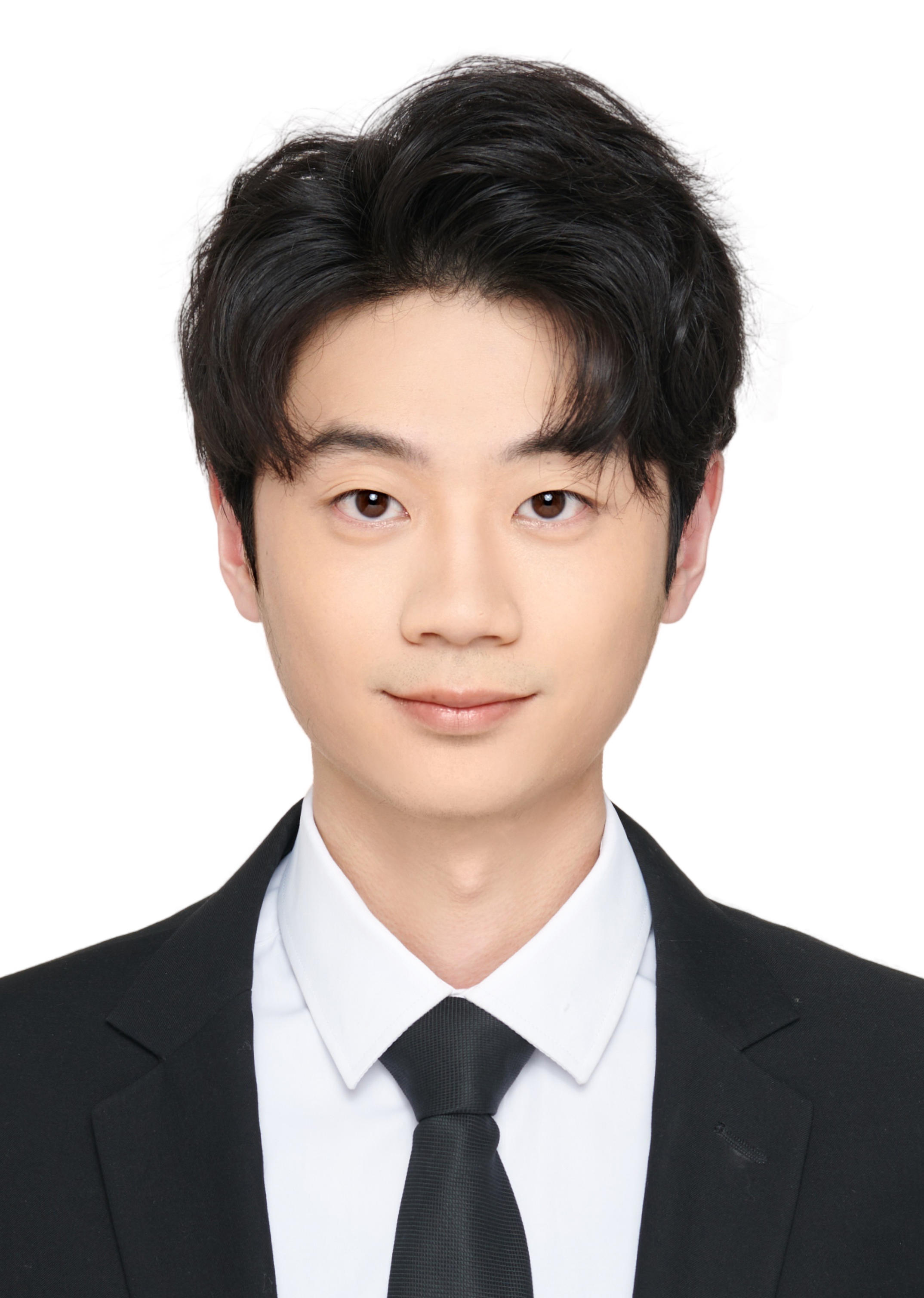}}]
	{Bin Wang} received the B.Sc. degree from the Chengdu University of Technology (CDUT), College of Computer Science and Cyber Security, Chengdu, China, in 2022, where he is currently pursuing the Ph.D. degree in software engineering with the College of Computer Science, Sichuan University, Chengdu. His research interests include intelligent geophysical data processing, remote sensing, computer vision, and domain adaptation.
\end{IEEEbiography}
\vspace{-12pt}
\begin{IEEEbiography}[{\includegraphics[width=1in,height=1in,clip,keepaspectratio]{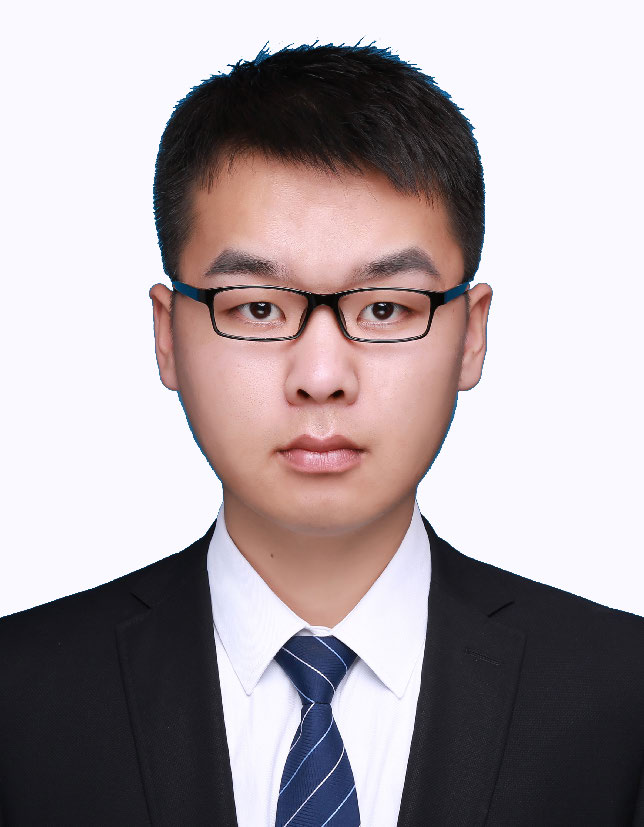}}]
	{Wenlong Gao}  received the B.S. degree from the Institute of Disaster Prevention, Langfang, Hebei, China, in 2018, and the M.S. degree from the School of Geophysics, Yangtze University, Wuhan, China,
	in 2021. He is currently pursuing the Ph.D. degree in geophysics with the School of Geophysics, Chengdu University of Technology, Chengdu, China. His research interests include forward modeling and inversion of frequency-domain controlled-source electromagnetic methods.
\end{IEEEbiography}

\bibliographystyle{IEEEtran}
		
\end{document}